\documentclass[10pt,twocolumn,letterpaper]{article}

\usepackage{iccv}
\usepackage{times}
\usepackage{epsfig}
\usepackage{graphicx}
\usepackage{amsmath}
\usepackage{amssymb}

\usepackage{subfigure}
\usepackage{multirow}
\usepackage{multicol}
\usepackage{amsthm}

\usepackage{mathrsfs}

\usepackage[breaklinks=true,bookmarks=false]{hyperref}

\iccvfinalcopy 

\begin{document}



\title{Reborn Mechanism: Rethinking the Negative Phase Information Flow in Convolutional Neural Network}

\author{Zhicheng Cai\\
School of Electronic Science and Engineering\\
Nanjing University\\
{\tt\small 181180002@smail.nju.edu.cn}
\and
Kaizhu Huang\\
School of Advanced Technology\\
Xi'an Jiaotong-liverpool University\\
{\tt\small kaizhu.huang@xjtlu.edu.cn}
\and
Chenglei Peng \thanks{Chenglei Peng is the corresponding author.}\\
School of Electronic Science and Engineering\\
Nanjing University\\
{\tt\small pcl@nju.edu.cn}
}

\maketitle

\begin{abstract}
   This paper proposes a novel nonlinear activation mechanism typically for convolutional neural network (CNN), named as \textbf{reborn mechanism}. In sharp contrast to ReLU which cuts off the negative phase value, the reborn mechanism enjoys the capacity to reborn and reconstruct dead neurons. Compared to other improved ReLU functions, reborn mechanism introduces a more proper way to utilize the negative phase information. Extensive experiments validate that this activation mechanism is able to enhance the model representation ability more significantly and make the better use of the input data information while maintaining the advantages of the original ReLU function. Moreover, reborn mechanism enables a non-symmetry that is hardly achieved by traditional CNNs and can act as a channel compensation method, offering competitive or even better performance but with fewer learned parameters than traditional methods. Reborn mechanism was tested on various benchmark datasets, all obtaining better performance than previous nonlinear activation functions.  
\end{abstract}
\section{Introduction}
\begin{figure}[htbp]
   \centering
   \subfigure{
      \includegraphics[width=0.43\textwidth]{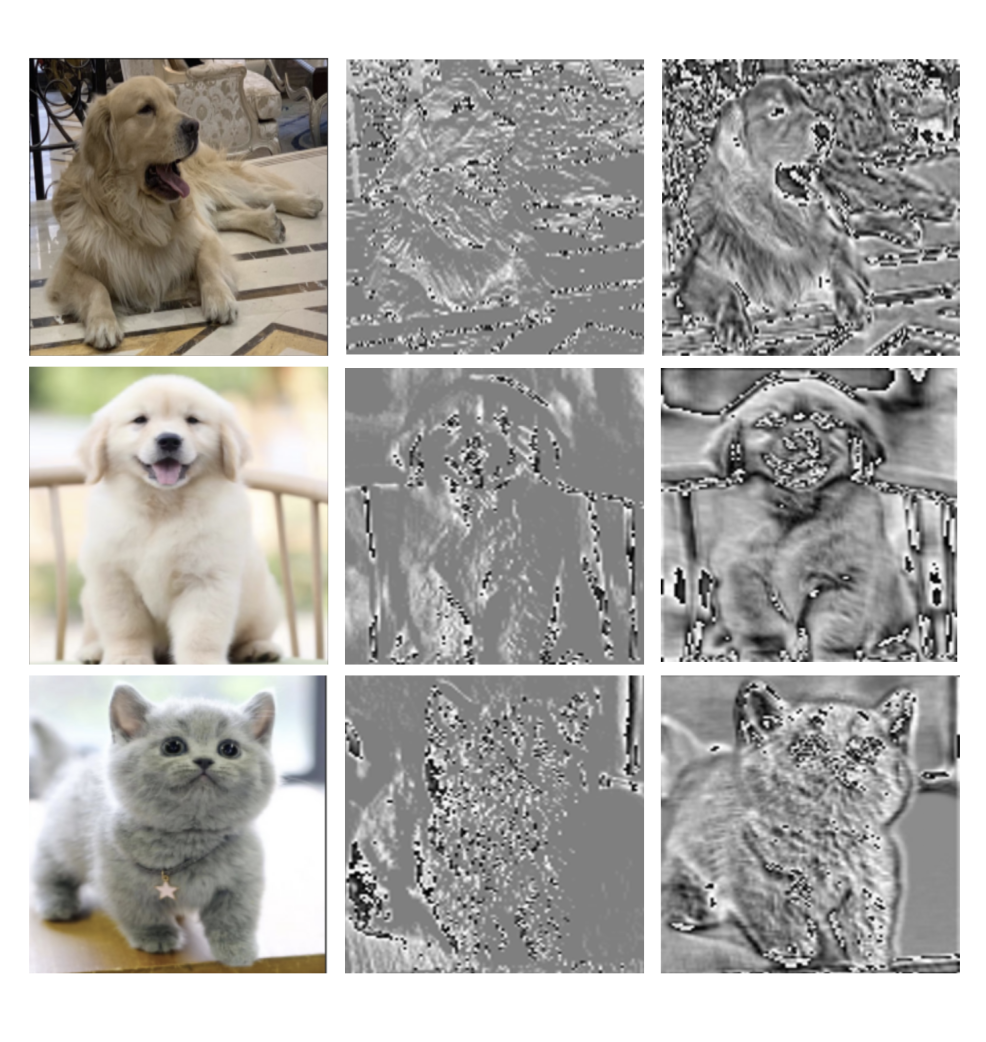}
      }
   \caption{Schematic diagram of the feature mappings activated by ReLU or reborn mechanism. \textbf{Left: }input images. \textbf{Middle: }feature mappings activated by ReLU. \textbf{Right: }feature mappings activated by reborn mechanism.  Feature mappings activated by reborn mechanism retain more valuable information and detect clearer outlines than those activated by ReLU.}
   \label{fff}
\end{figure}

\begin{figure*}[htbp]
   \centering
   \subfigure{
      \includegraphics[width=0.43\textwidth]{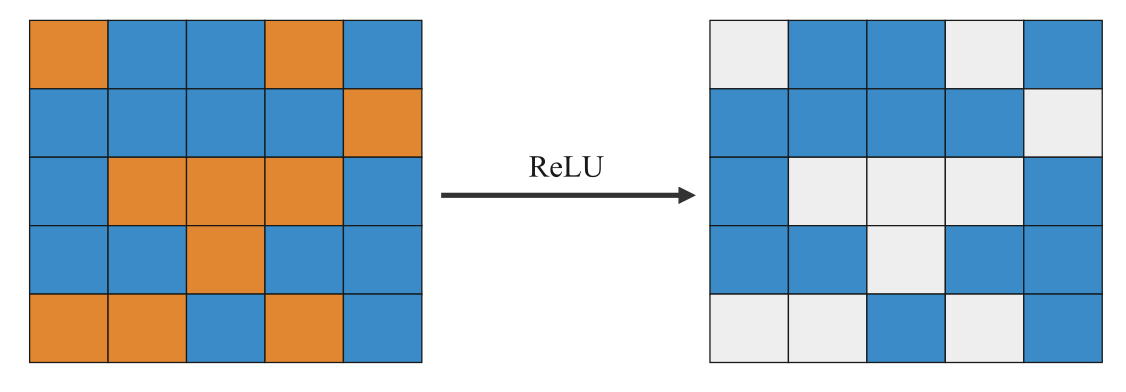}
   }
   \subfigure{
      \includegraphics[width=0.43\textwidth]{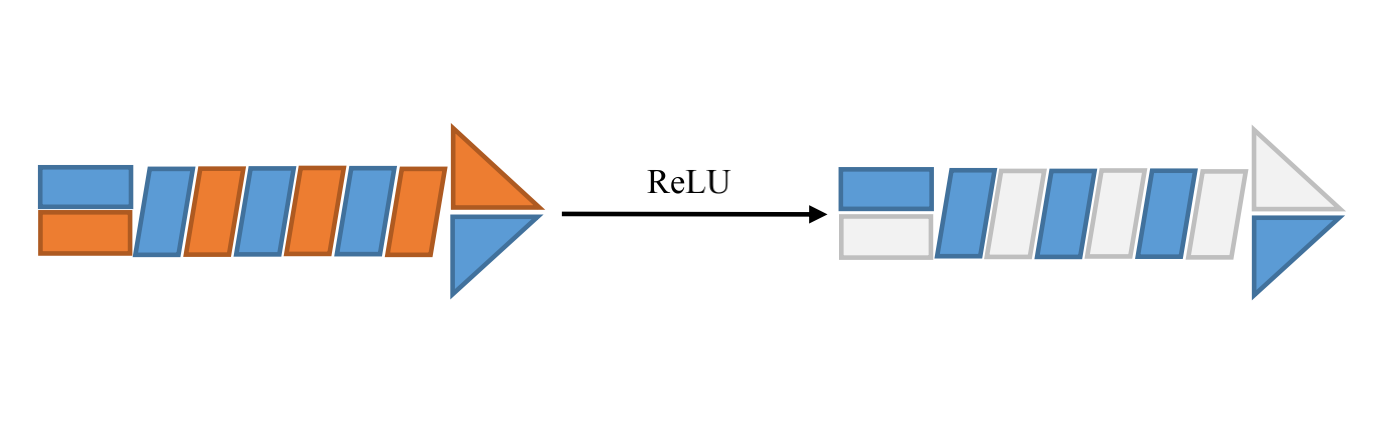}
   }
   \subfigure{
      \includegraphics[width=0.43\textwidth]{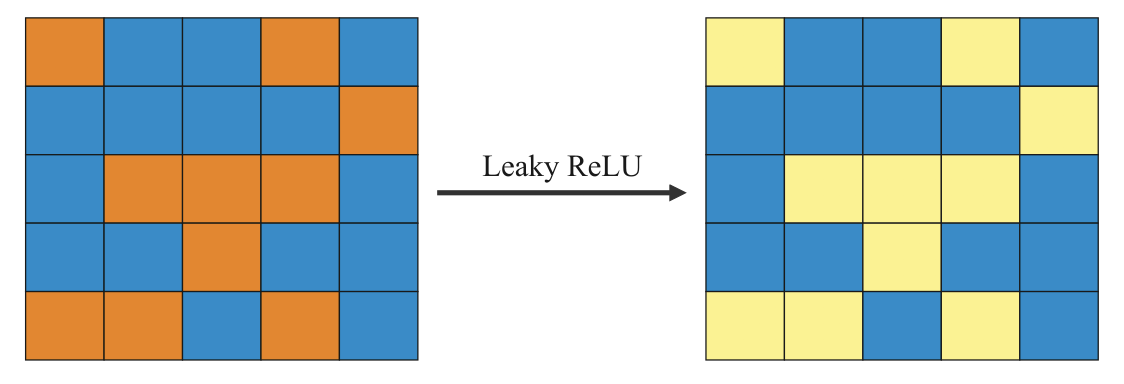}
   }
   \subfigure{
      \includegraphics[width=0.43\textwidth]{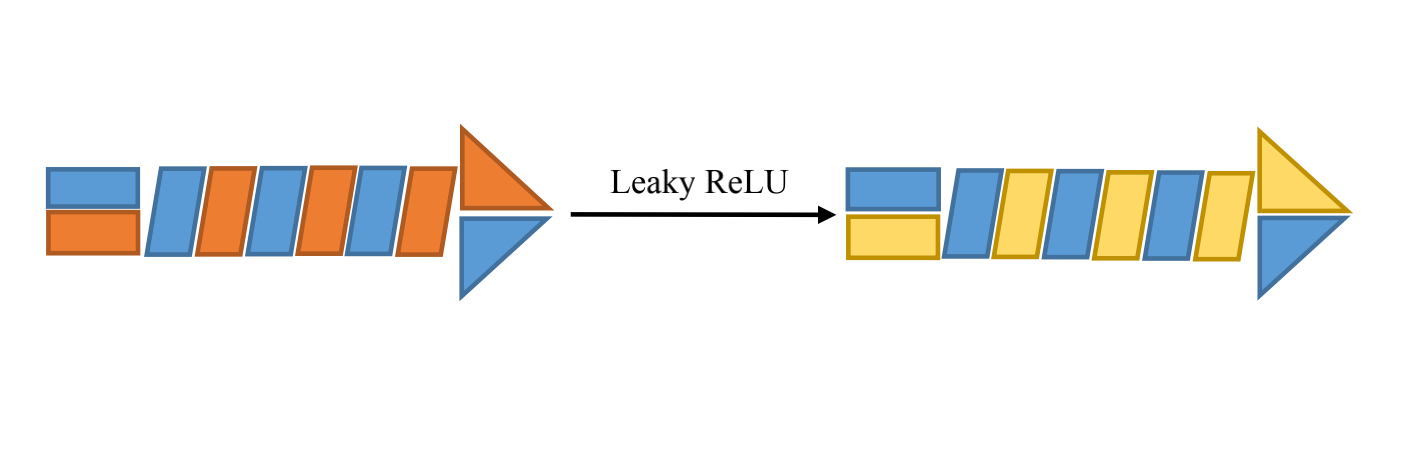}
   }
   \subfigure{
      \includegraphics[width=0.43\textwidth]{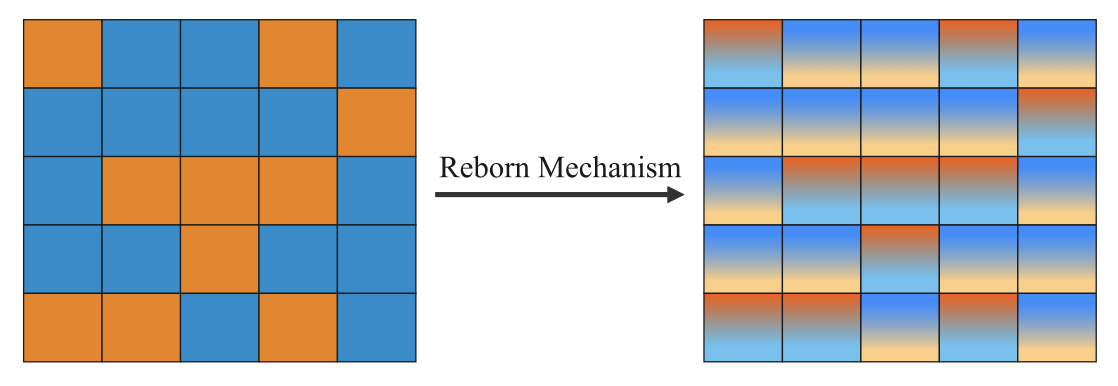}
   }
   \subfigure{
      \includegraphics[width=0.43\textwidth]{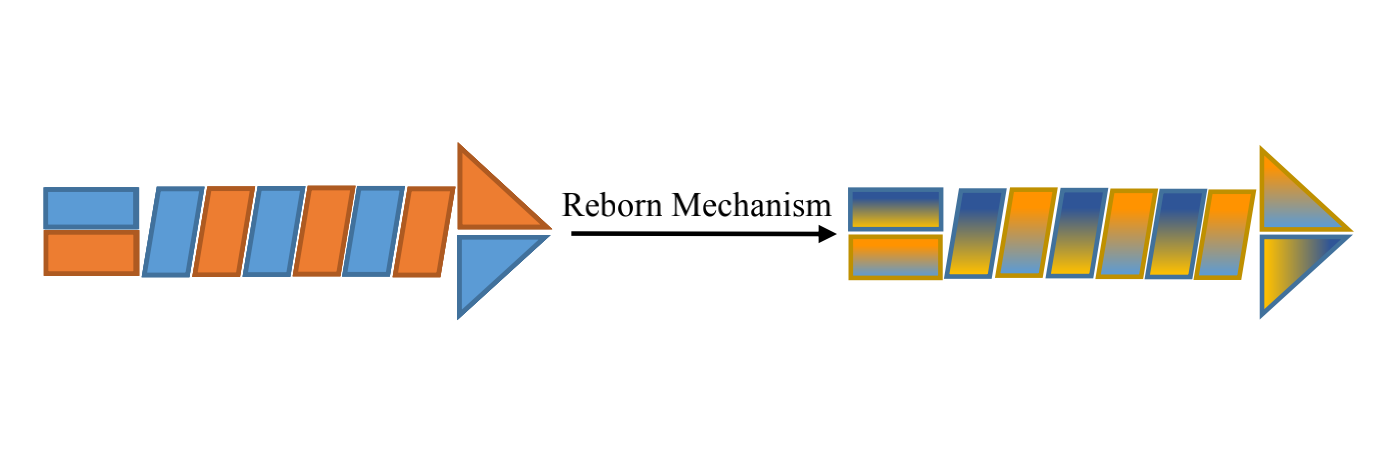}
   }
   \caption{Schematic diagram of the activation procedures conducted by ReLU, Leaky ReLU and Reborn Mechanism. \textbf{Top:} activation procedure of ReLU. \textbf{Middle:} activation procedure of improved ReLUs (e.g. Leaky ReLU). \textbf{Bottom:} activation procedure of Reborn Mechanism. \textbf{Left:} exhibit three activation procedures in the format of feature mapping. \textbf{Right:} exhibit three activation procedures in the format of information flow. Blue parts mean positive values, orange parts represent negative information, gray parts stand for zero values. Yellow parts indicate the negative information which has been multiplied with light weights and become less significant, the parts whose color changes gradually stand for the negative values reconstructed by reborn mechanism concatenated with positive values. The way that reborn mechanism processes the negative information is more proper compared to the simple truncation by ReLU and light weight multiplication by Leaky ReLU.}
   \label{f}
\end{figure*}

Since AlexNet~\cite{krizhevsky2012imagenet} won the ILSVRC-2012, convolutional neural network (CNN) has flourished in a compelling way~\cite{szegedy2015going,he2016deep,howard2019searching} and been successfully applied in many real scenarios~\cite{alp2018densepose,zhang2017stackgan,qi2017pointnet}.

The feature map or the effective use of input information at all layers is believed to be vital for CNN models. However, it is still an outstanding problem: how can CNN models make full and effective use of information and reduce redundant features? It is known that the CNN typically adopts the rectifier linear unit (ReLU)~\cite{glorot2010deep} as the activation function. Although ReLU guarantees the nonlinearity and brings the advantages of sparsity, the forward propagation process will truncate and discard all neurons with negative values. Moreover, the back-propagation will cause the death of a certain number of neurons. As a result, the data information flow with negative phase is completely obstructed through the network and the information contained in these truncated neurons fails to adapt fully and effectively. Furthermore, the  information loss will accumulate through convolutional layers. While some improved ReLUs like Leaky ReLU~\cite{maas2013rectifier} take advantage of negative phase information, they lose some of the ReLU's original advantages. In this paper, we argue that these methods may not tackle the negative phase information properly in that they simply multiply the negative values with relatively light weights. As shown later in the experiments, replacing ReLU with these improved ReLUs fails to improve the model's ability significantly.

Our work raises a novel nonlinear activation mechanism named \textbf{reborn mechanism} to implement nonlinear activation operation in CNN models. The \textbf{reborn block} is introduced to realize the reborn mechanism. The core of reborn block is utilizing deconvolution operation to reconstruct the neurons cut off by ReLU in the activation layer. Consequently, the information carried by the reborn neurons are fused to the forward propagation by  channel concatenation with the positive feature mappings. As such, the negative phase information can be processed more properly. Fig.~\ref{fff} exhibits the feature mappings activated by ReLU or reborn mechanism, which will be discussed specifically later. Fig.~\ref{f} illustrates three different activation methods on how to process the negative phase information.  In summary, our activation mechanism aims to take full advantage of the information carried by all neurons, utilize the negative phase information properly, and reduce the loss of information in the network flow without sacrificing the benefits that ReLU provides. 

Except for the effective utilization of information, reborn block possesses many other advantages. First, the usage of negative information indicates that reborn mechanism can act as a method of channel compensation~\cite{shang2016understanding}. As a result, reborn mechanism can enhance the utilization effectiveness of convolutional kernels and achieve the same performance with fewer convolutional kernels. Second, the unsymmetrical architecture of the reborn block can break the symmetry of CNN models, which improves the model representation ability. Moreover, reborn block has a nice portability and can be utilized in any CNN backbone models. In addition, we argue that reborn mechanism is more than a nonlinear activation function typified by ReLU, to be specific, it is a nonlinear activation mechanism which possesses certain architecture and conducts the nonlinear activation operation. The advantages of reborn mechanism will be detailed in the following sections.


\section{Related Work}
\subsection{Nonlinear Activation Functions}
The nonlinear activation function follows a weight layer to conduct nonlinear operation, allowing neural networks to learn the nonlinear mapping. A common and effective nonlinear activation function is the rectified linear unit (ReLU) function~\cite{glorot2010deep}, as shown in Eq.~\ref{eq1}.
\begin{equation}
f_{(ReLU)}(x)=max(0,x)\label{eq1}
\end{equation}
ReLU solves the problem about gradient and increases the network's sparsity. However, if the gradient of the neuron is zero, the neuron's weight will not be updated, leading to the death of the neuron. Moreover, it is believed that the negative neurons hold valuable information as well, but ReLU carries out simple truncation and discards information from negative neurons, resulting in the low effective utilization of the input information. To this end, some researchers improve the ReLU function by taking advantage of negative values, such as LeakyReLU~\cite{maas2013rectifier}, PReLU~\cite{he2015delving} and so on \cite{clevert2015fast,klambauer2017self,hendrycks2016gaussian}. 

Although these improved ReLUs utilize the negative phase information by multiplying them with light weights, they inevitably lose some advantages of ReLU. There also exist empirical investigations \cite{saxe2013exact} showing that under certain circumstance, the improved ReLUs do not exert a better effect than ReLU. As a matter of fact, we argue that the way these improved ReLUs make use of the negative phase information may not be proper in that the negative phase information multiplied with light weights is less effectively explored. To alleviate this problem, we design an activation mechanism which is expected to make better use of the input data information and enhance model representation ability.

\subsection{Deconvolution Operation}
Deconvolution \cite{zeiler2010deconvolutional} is a type of up-sampling method, which is an essential component in semantic segmentation~\cite{noh2015learning} and generative adversarial networks~\cite{radford2015unsupervised}. Such operation is also used to visualize CNN in ZFNet~\cite{zeiler2014visualizing}. Contrary to the many-to-one mapping of convolution operation, deconvolution focuses on the set of one-to-many mapping. As a result, it can be considered as the reverse operation of convolution. 

In reborn block, deconvolution operation is the key element that is employed to process the negative phase information such that the negative neurons dead otherwise can be regenerated and reconstructed. After the deconvolution operation, the negative phase information will flow forward with the positive phase information by concatenating and compressing these two feature maps. 

\section{Reborn Mechanism}
\begin{figure}[htbp]
\centering
\includegraphics[height=0.56\textwidth]{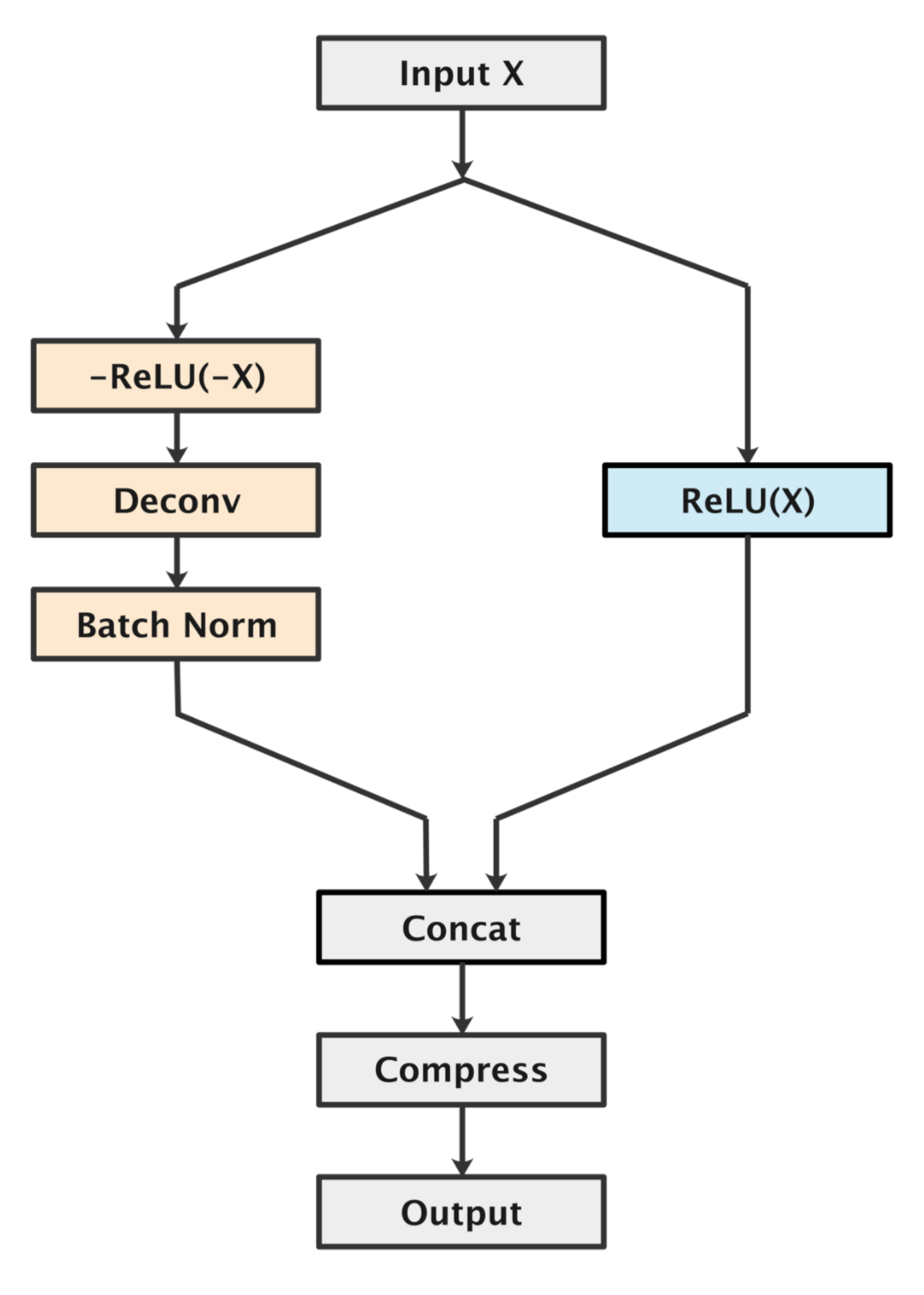}
\caption{Reborn block}
\label{fig2}
\end{figure}
\subsection{Reborn Mechanism and Reborn Block}
According to the above analysis, although ReLU activation function can increase the sparsity of the network and solve the problems related to the gradient, it truncates all negative values and zeros the neuron value, leading that some neurons perish and consequently useful information disappears. This naturally raises one question that how we can make full use of all the input information without sacrificing  the advantages of ReLU. While arguing that the information with negative phase is also essential, we rethink the flow mode of negative phase information in the deep CNN models. We introduce a novel nonlinear activation mechanism to activate neurons and process negative phase information. In this way, negative neurons perished previously can be reborn and reconstructed, therefore information in the negative phase is able to propagate forward and backward effectively. This mechanism is called the ``reborn mechanism''. The module that introduces and implements the reborn mechanism is termed as reborn block.

We now turn to discuss the operation conducted by the reborn block in details. Suppose that the reborn block's input $X$ is the feature map obtained from the upper convolutional layer. Firstly, $X$ is input into ReLU to obtain the activated feature map $X_1$. Then the positive neurons are selected and the negative ones are truncated. Simultaneously, $-X$ is input into another paralleled ReLU function to obtain the activated feature map $X_2^*$ followed by screening out the negative values neurons and truncating the positive neurons. In order to retain the phase of the gradient, the activated feature map $X_2^*$ is inverted. Then $-X_2^*$ is deconvoluted and batch normalized to obtain the feature map $X_2$. After that, these two feature maps $X_1$ and $X_2$ are concatenated together. To compress the doubled feature channels, one $1\times1$ convolutional layer is employed after the concatenation operation. Moreover, no activation functions are applied between the $1\times1$ convolutional layer and the next convolutional layer or fully connected layer. Now we obtain the feature map receiving the nonlinear activation operation. Fig.~\ref{fig2} shows the schematic diagram of the reborn block. Obviously, the operation that the reborn block conducts is nonlinear, and this is more than a function which cannot be simply written in the formation of normal function expression and drawn the function curve. As a result, we consider this operation as a kind of mechanism, and we call this mechanism as ``reborn mechanism''.

For clarity, we define some notations in Table~\ref{lon}.
\begin{table}[htbp]
   \begin{center}
   \begin{tabular}{c|l}
   \hline
    symbol & \ \ \ \ \ \ \ \ \ \ \ \ \ \ \ \ \ \ \ \ meaning  \\
    \hline
    $\sigma(x)$ & ReLU function \\
    \hline
    $\Phi(x)$ & Deconvolution operation \\
    \hline
    $\kappa(x)$ & Batch normalization operation \\
    \hline
    $\Psi(x_1,x_2,\cdot \cdot \cdot,x_n)$ & Channel concatenating operation \\
    \hline
    $\Gamma(x,n)$ & Channel compression operation\\
    \hline
    $n$ & Decay ratio of channel number \\
    
   \hline
   \end{tabular}
   \end{center}
   \caption{List of Notation}
   \label{lon}
   \end{table}

Thus, the reborn block can be mathematically formulated as Eq.~\ref{eq3}:
\begin{equation}
f(X) = \Gamma(\Psi(\sigma(X),\kappa(\Phi(-\sigma(-X)))),2)\label{eq3}
\end{equation}
\subsection{Interpretations for Reborn Mechanism}
In this section, we will interpret the idea of reborn mechanism and reborn block. Intuitively, in reborn block, those negative neurons exceeding a certain threshold are incapable of retaining the values they carry and passing them forward. In this case, they will be given a chance to return to their previous state and perform convolution again. As a result, certain reborn neurons are enabled to retain their information in this samsara. This process is equivalent to performing a second screening. However, such rebirth process may not be free, and the reborn neurons shall have weaker importance and smaller weight than the positive neurons that have already met the conditions. Therefore, the used deconvolution represents ``an opportunity of rebirth'', also as ``a cost of rebirth''. In this regard, though the information with negative phase is not as important as that with the positive phase, a more proper way to process the negative information is designed that can filter them for the second chance to avoid discarding valuable information.

We can also interpret the reborn block in a way that the negative information fails to meet the threshold condition and therefore cannot activate the neuron as is done to the positive information. However, they still carry useful information. Since they cannot simultaneously flow forward in parallel, they are forced to be lagged, meaning that the negative flow is pushed back one level. The deconvolution introduces a hysteresis cost as well as the implementation of hysteresis. Deconvolution is equivalent to decouple these negative values, and finally, it is the decoupled filtered information that is still active. Additionally, the operation of deconvolution and channel concatenation is a kind of deep and shallow feature fusion, concerning only the feature with negative phase.

\subsection{Advantages of Reborn Mechanism}

\textbf{Effective Usage of Input Information:} We assume reasonably that features with negative phase possesses valuable information as well. In this sense, it is inappropriate to simply discard the negative phase information, as done in the traditional ReLU. Our reborn mechanism offers a more proper way to process and make use of the negative phase information, which enables the CNN model to take full advantage of the input data, and reduce the loss of data information. Moreover, this kind of nonlinear activation mechanism does not lose the original advantages of ReLU, remains the nonlinear of the model, enhances the sparsity of the network weight matrices, and avoids the problems of gradient explosion and gradient varnishing.

\textbf{Unsymmetrical Network Design:} In addition, with the reborn block, the network can be unnecessarily symmetrical, which is beneficial to enhancing the network's representation ability. Suppose that only a small number of neurons change their activated values according to the different input value, and the majority of the neurons are insensitive to the different input values, the order of the weight matrix is small, and the total order will be significantly smaller after the multiplication of weight matrices. As a result, although the dimension of weight matrices is high, most of the dimensions possess little valuable information, weakening the representation ability of the model. Such network degradation problem is partially owed to symmetrical characteristic of the network. However, the reborn block employs an architecture with two parallel and unsymmetrical paths which process negative phase and positive phase information differently and simultaneously. It breaks the symmetrical characteristic of CNN network, and enhances the representation ability of the model as a result.

\textbf{Channel Compensation:} Our reborn block can also be considered as a channel compensation method that enhances the efficiency of feature channels. Shang et al. pointed that at the lower layers of a CNN model, the network is sensitive to both positive phase information and negative phase information, tending to capture these two kinds of information simultaneously~\cite{shang2016understanding}. However, the ReLU function forbids the negative phase information to pass, leading to the redundancy of convolutional kernels. Therefore, the model needs more convolutional kernels. This problem is called ``Complementary phenomenon of network parameters'' ~\cite{shang2016understanding}. The reborn block can alleviate this problem by inverting the input channels simultaneously, processing the negative phase information with deconvolution operation and finally concatenating these two feature maps. As a result, the feature channels in the model with reborn block are used more effectively with a fewer number of convolutional kernels to achieve the same performance as that of the model with ReLU, which is verified in latter experiments. Compared to the CReLU \cite{shang2016understanding} which uses both negative and positive information with simple concatenation, our reborn mechanism processes the negative phase information more properly and conceivably. Our experiments also validates this point as later seen in Section 4.

\textbf{Portability:} As a kind of activation mechanism supersetting ordinary activation function, reborn mechanism can be used in any CNN models, indicating that it has the advantage of portability. Moreover, in the reborn block, the utilization of the $1\times1$ convolutional layer is only aimed to compress the number of channels to the original level, making it easier to be implemented in other CNN models without fine-tuning any feature channels. Our experimental results showed that if we discard the $1\times1$ convolutional layer and double the input channel number of the next $3\times3$ convolutional layer, the result would only have negligible change. In addition, adjusting the kernel size from $1\times1$ to $3\times3$ makes no difference. It is also noted that,  the reborn mechanism can be readily applied in other machine learning models such as MLP and RNN. We will leave this investigation as future work. 

\section{Experiment}
\begin{table*}[htbp]
   \begin{center}
   \begin{tabular}{|l|c|c|c|c|c|c|}
   \hline
   Method & CIFAR-10 & CIFAR-100 & SVHN & MNIST & KMNIST & FMNIST \\
   \hline
   \multicolumn{7}{|c|}{ConvNet-8}\\
   \hline
   ReLU & 90.39\%& 67.27\% & 93.96\%& 99.50\% &97.06\%  &92.95\%\\
   \hline
   CReLU & 89.83\% & 66.71\%&  94.37\% & 99.41\% & 97.68\% & 93.02\%\\
   \hline
   Leaky ReLU & 90.43\% & 66.89\% & 93.76\% & 99.42\% &96.64\% &93.09\%\\
   \hline
   PReLU & 89.38\%&66.30\% & 93.64\%& 99.52\% &96.97\% &92.84\%\\
   \hline
   RReLU & 90.66\%& 68.09\% & 94.05\%& 99.41\% &97.09\% &93.20\%\\
   \hline
   CELU & 89.45\%& 65.62\% & 93.57\%& 99.41\% &96.59\% &92.76\%\\
   \hline
   SELU & 87.92\%& 61.13\% & 93.50\%& 99.41\% &96.25\%&92.65\%\\
   \hline
   ELU & 90.00\% & 65.92\% & 93.42\%& 99.41\% &96.24\%&92.82\%\\
   \hline
   Reborn Mechanism & \textbf{91.80\%} & \textbf{68.41\%} & \textbf{95.34\%} & \textbf{99.59\%} &\textbf{98.24\%} &\textbf{93.66\%}\\   
   \hline
   \multicolumn{7}{|c|}{ConvNet-13}\\
   \hline
   ReLU & 91.58\%& 68.48\% & 94.82\%& 99.55\% &97.55\%  &93.36\%\\
   \hline
   CReLU & 92.06\% & 69.07\%&  95.37\% & 99.61\% &97.78\% & 93.39\%\\
   \hline
   Leaky ReLU & 91.52\% & 68.89\% & 94.86\% & 99.51\% &97.47\% &93.52\%\\
   \hline
   PReLU & 91.08\%&68.11\% & 94.55\%& 99.50\% &97.55\% &93.71\%\\
   \hline
   RReLU & 91.75\%& 69.38\% & 94.74\%& 99.48\% &97.69\% &93.48\%\\
   \hline
   CELU & 89.55\%& 68.67\% & 94.57\%& 99.47\% &97.43\% &92.793\%\\
   \hline
   SELU & 87.61\%& 65.02\% & 94.27\%& 99.41\% &97.20\%&92.71\%\\
   \hline
   ELU & 90.22\% & 66.34\% & 94.31\%& 99.43\% &97.39\%&92.92\%\\
   \hline
   Reborn Mechanism & \textbf{93.17\%} & \textbf{70.77\%} & \textbf{95.81\%} & \textbf{99.70\%} &\textbf{98.51\%} &\textbf{93.86\%}\\   
   \hline
   \multicolumn{7}{|c|}{ResNet-34}\\
   \hline
   ReLU & 93.64\%& 74.89\% & 95.43\%& 99.57\% &98.17\%  &94.23\%\\
   \hline
   CReLU & 93.89\% & 75.32\%&  95.67\% & 99.61\% &98.34\% & 94.17\%\\
   \hline
   Leaky ReLU & 93.56\% & 74.37\% & 95.41\% & 99.51\% &98.17\% &94.17\%\\
   \hline
   PReLU & 93.46\%&72.68\% & 95.45\%& 99.50\% &97.95\% &94.33\%\\
   \hline
   RReLU & 93.77\%& 74.91\% & 95.53\%& 99.48\% &98.09\% &94.35\%\\
   \hline
   CELU & 92.52\%& 72.34\% & 95.47\%& 99.47\% &97.63\% &93.76\%\\
   \hline
   SELU & 90.64\%& 70.22\% & 94.93\%& 99.41\% &97.40\%&93.88\%\\
   \hline
   ELU & 92.71\% & 72.86\% & 95.01\%& 99.43\% &97.49\%&93.91\%\\
   \hline
   Reborn Mechanism & \textbf{94.73\%} & \textbf{76.56\%} & \textbf{96.72\%} & \textbf{99.73\%} &\textbf{99.05\%} &\textbf{95.01\%}\\   
   \hline
   \end{tabular}
   \end{center}
   \caption{Test accuracy of different methods based on three basic CNN models on various benchmark datasets}
   \label{tab1}
   \end{table*}

\subsection{Dataset}
In our experiments, the CNN models with reborn mechanism or other traditional activation functions were trained and tested on six benchmark datasets, namely, CIFAR-10, CIFAR-100, SVHN, MNIST, KMNIST, and Fashion-MNIST.

\textbf{CIFAR-10:} CIFAR-10 dataset is composed of 10 classes of natural images with 50,000 training images in total, and 10,000 testing images. The 10 classes include: airplane, automobile, bird, cat, deer, dog, frog, horse, ship, and truck. Each image is a RGB image of size $32\times32$. 

\textbf{CIFAR-100:} CIFAR-100 dataset is composed of 100 different classifications, and each classification includes 600 different color images, of which 500 are training images and 100 are test images. As a matter of fact, these 100 classes are composed of 20 super classes, and each super class possesses 5 child classes. The images in the CIFAR-100 dataset have a size of $32\times32$ like CIFAR-10. 

\textbf{SVHN:} SVHN (Street View House Numbers) dataset is composed of 630,420 RGB digital images with a size of $32\times32$, including a training set with 73,257 images and a test set with 26,032 images. We did not use the extra set.

\textbf{MNIST:} MNIST is a 10-class dataset of 0$\sim$9 handwritten digits, including 60,000 training images and 10,000 test images in total. Each sample is a size $28\times28$ gray-scale image. 

\textbf{KMNIST:} Kuzushiji-MNIST (KMNIST) dataset is composed of 10 classes of cursive Japanese characters (namely, ``Kuzushiji'').  Each sample is a gray-scale image of $28\times28$ size. This dataset includes 60,000 training images and 10,000 test images in total. 

\textbf{Fashion-MNIST:} Fashion-MNIST (FMNIST for short) is a 10-class dataset of fashion items: T-shirt/top, Trouser, Pullover, Dress, Coat, Scandal, Shirt, Sneaker, Bag and Ankle boot. Each sample is a gray-scale image of $28\times28$ size. This dataset includes 60,000 training images and 10,000 test images in total.

\subsection{Basic CNN Models}

To make our experimental results more cogent, we not only used various benchmark datasets, but also applied three different classic CNN models with different depths, widths and architectures as the basic models, namely, ConvNet-8, ConvNet-13, and ResNet-34.

\textbf{ConvNet-8:} ConvNet-8 is a relatively small CNN model composed of eight $3\times3$ convolutional layers and three fully connected layers. The first convolutional layer has $32$ convolutional kernels, and the number of convolutional kernels doubles every two convolutional layers.

\textbf{ConvNet-13:} ConvNet-13 has the same configuration of convolutional layers as VGG-16 Net\cite{simonyan2014very}, but we used convolution with step-size instead of the maximum pooling, replaced the three fully connected layers with the global average pooling~\cite{cai2021study}, and added batch normalization layers after these thirteen convolutional layers. 

\textbf{ResNet-34:} ResNet-34 has exactly the same architecture as stated in \cite{he2016deep} which is characterized with residual connection, and employs step-size convolution instead of pooling.

\begin{table*}[htbp]
\begin{center}
\begin{tabular}{|l|c|c|c|c|c|c|}
\hline
Method & CIFAR-10 & CIFAR-100 & SVHN & MNIST & KMNIST & FMNIST \\
\hline
\multicolumn{7}{|c|}{ConvNet-8}\\
\hline
ReLU & 90.39\%& 67.27\% & \textbf{93.96\%}& 99.50\% &97.06\%  &92.95\%\\
\hline
CReLU & 89.83\% & 66.71\%&  94.37\% & 99.41\% & \textbf{97.68\%} & 93.02\%\\
\hline
Half ReLU & 88.91\%& 65.22\% & 92.91\%& 99.43\% &96.11\%  &92.45\%\\
\hline
Half CReLU & 88.62 \% & 65.01\%&  93.56\% & 99.36\% & 97.15\% & 92.93\%\\
\hline
Half Reborn Mechanism & \textbf{91.19\%}& \textbf{67.35}\% & 93.90\%& \textbf{99.53\%} &96.92\%  &\textbf{93.05\%}\\
\hline
\multicolumn{7}{|c|}{ConvNet-13}\\
\hline
ReLU & 91.58\%& 68.48\% & 94.82\%& 99.55\% &97.55\%  &93.36\%\\
\hline
CReLU & \textbf{92.06\%} & \textbf{69.07\%}&  95.37\% & 99.61\% &97.78\% & \textbf{93.39\%}\\
\hline
Half ReLU & 90.45\%& 67.25\% & 94.67\%& 99.42\% &97.56\%  &92.56\%\\
\hline
Half CReLU & 90.56\%& 67.10\% & 95.11\%& 99.51\% &97.61\%  &92.74\%\\
\hline
Half Reborn Mechanism & 91.74\%& 68.56\% & \textbf{95.56\%}& \textbf{99.63\%} &\textbf{98.27\%}  &\textbf{93.39\%}\\
\hline
\multicolumn{7}{|c|}{ResNet-34}\\
\hline
ReLU & 93.64\%& 74.89\% & 95.43\%& 99.57\% &98.17\%  &\textbf{94.23\%}\\
\hline
CReLU & \textbf{93.89\%} & \textbf{75.32\%}&  95.67\% & 99.61\% &\textbf{98.34\%} & 94.17\%\\
\hline
Half ReLU & 92.65\%& 72.66\% & 94.23\%& 99.56\% &97.87\%  &93.57\%\\
\hline
Half CReLU & 92.76 \% & 73.14\%&  95.21\% & 99.54\% &98.04\% & 93.91\%\\
\hline
Half Reborn Mechanism & 93.15\%& 75.23\% & \textbf{95.73\%}& \textbf{99.62\%} &98.22\%  &93.97\%\\
\hline
\end{tabular}
\end{center}
\caption{Comparison of various methods with different network width.  Reborn mechanism enjoys the advantage of channel compensation.}
\label{tab2}
\end{table*}
\begin{table*}[htbp]
\begin{center}
\begin{tabular}{|l|c|c|c|c|c|c|}
\hline
Method & CIFAR-10 & CIFAR-100 & SVHN & MNIST & KMNIST & FMNIST \\
\hline
\multicolumn{7}{|c|}{ConvNet-8}\\
\hline
ReLU & 90.39\%& 67.27\% & 93.96\%& 99.50\% &97.06\%  &92.95\%\\
\hline
Extended Conv & 90.45\%& 67.32\% & 93.93\%& 99.54\% &96.92\%  &93.17\%\\
\hline
Reborn Mechanism & \textbf{91.80\%} & \textbf{68.41\%} & \textbf{95.34\%} & \textbf{99.59\%} &\textbf{98.24\%} &\textbf{93.66\%}\\   
\hline
\multicolumn{7}{|c|}{ConvNet-13}\\
\hline
ReLU & 91.58\%& 68.48\% & 94.82\%& 99.55\% &97.55\%  &93.36\%\\
\hline
Extended Conv & 90.54\%& 66.79\% & 94.96\%& 99.54\% &97.75\%  &93.09\%\\
\hline
Reborn Mechanism & \textbf{93.17\%} & \textbf{70.77\%} & \textbf{95.81\%} & \textbf{99.70\%} &\textbf{98.51\%} &\textbf{93.86\%} \\   
\hline
\multicolumn{7}{|c|}{ResNet-34}\\
\hline
ReLU & 93.64\%& 74.89\% & 95.43\%& 99.57\% &98.17\%  &94.23\%\\
\hline
Extended Conv & 93.67\%& 74.93\% & 95.36\%& 99.57\% &98.21\%  &94.25\%\\
\hline
Reborn Mechanism & \textbf{94.73\%} & \textbf{76.56\%} & \textbf{96.72\%} & \textbf{99.73\% }&\textbf{99.05\%} &\textbf{95.01\%}\\   
\hline
\end{tabular}
\end{center}
\caption{Test accuracy of three CNN models with reborn blocks, extended convolutional layers, and ReLU on various benchmark datasets.}
\label{tab3}
\end{table*}
\subsection{Training Details}
In all experiments, we adopted the SGD algorithm with a batch size of $64$ and the hyper-parameter of the momentum  is set to $0.9$. The method of weight decay was applied to all models, and the ratio of weight decay is empirically set as $0.0005$. For CIFAR-10 and CIFAR-100, we trained all the models $160$ epochs, and for the rest datasets, the number of epochs was $80$. In addition, the hyper-parameter learning rate was $0.001$ in the first half epochs and then changed to $0.0001$ in the later half epochs. Moreover, in the experiments of CIFAR10, CIFAR100, and FMNIST, we exploited the data augmentation that first padded 4 circles of 0 pixels around the input image, and then randomly cut it to the original size, then flipped the image horizontally with a probability of $50\%$. We did not apply any data augmentation method for the rest three datasets. All the learned parameters in these models are initialized by Xavier initialization method.

To verify that the reborn block improves the model's performance indeed, we conducted comparative experiments where the basic CNN models were equipped with different nonlinear activation methods. Specifically, we kept all of the hyper-parameters unchanged and simply replaced the traditional ReLU function in the basic CNN models with reborn mechanism or other improved activation functions including CReLU, Leaky ReLU, PReLU, RReLU, CELU, SELU, and ELU. After that, we employed the same training methods to train these CNN models on six benchmark datasets. Finally, we compared these test results of the basic CNN models with reborn blocks and other traditional activation functions. We used the NVIDIA GeForce RTX 2080 Ti GPU to train all the models.

\begin{figure*}[htbp]
   \centering
   \subfigure{
      \includegraphics[width=0.82\textwidth]{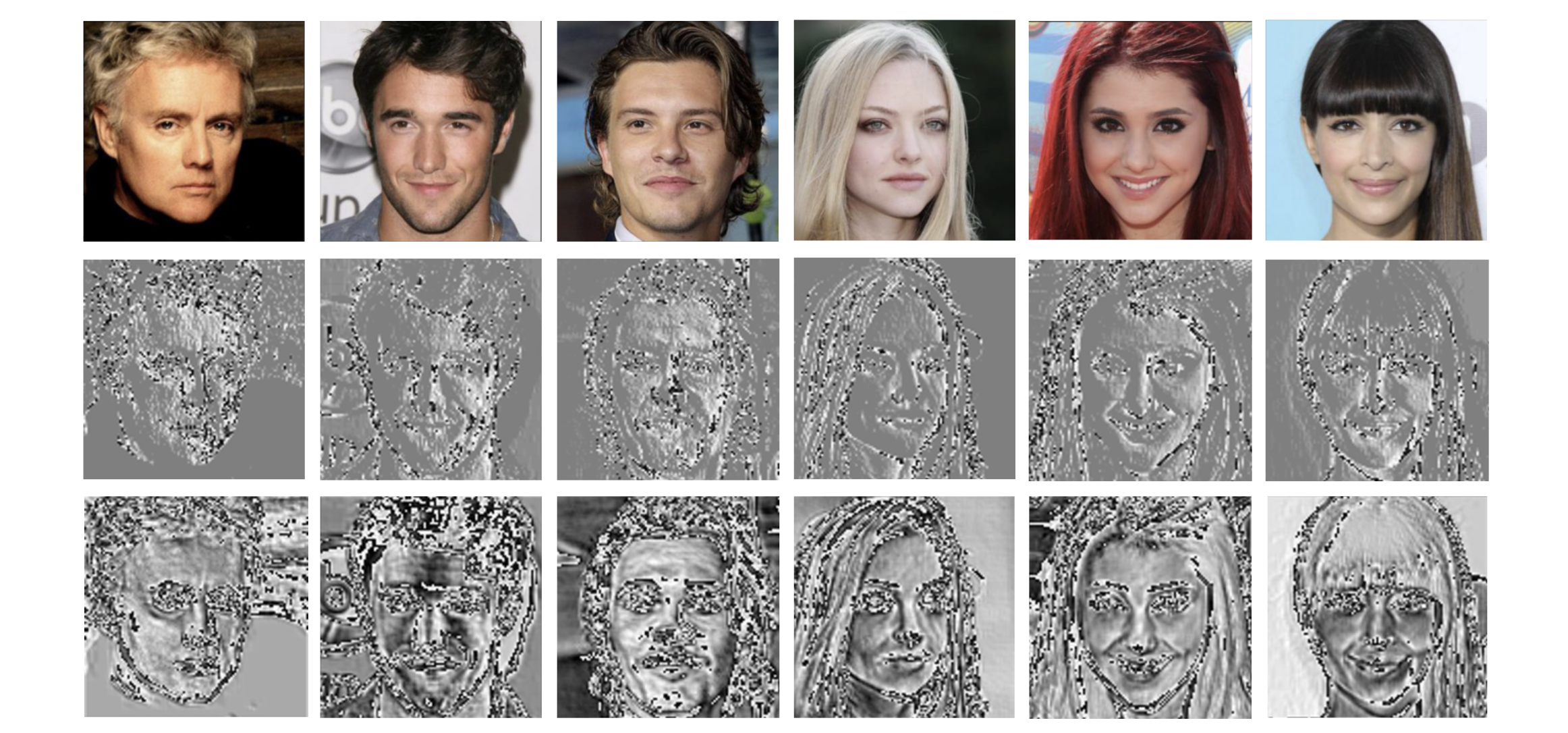}
   }
   \caption{Visualization of various feature mappings activated by reborn mechanism and ReLU. \textbf{Top: }the input images. \textbf{Middle: }the feature mappings activated by ReLU. \textbf{Bottom: }the feature mappings activated by reborn mechanism. The white pixels possess zero values. More gray pixels possess larger values. It is clearly exhibited that feature mappings activated by reborn mechanism retain more valuable information and detect clearer outlines than those activated by ReLU. In addition, reborn mechanism detects more details like eyes, mouths, and noses, which ReLU fails to. (Images here are selected from CelebA-HQ dataset~\cite{karras2017Progressive}) }
   \label{pic}
\end{figure*}
\subsection{Experimental Results}
Table~\ref{tab1} shows all the experimental results, i.e. the test accuracies obtained by the different methods on various benchmark datasets. It is clearly shown that reborn activation mechanism enabled the CNN models to obtain significantly better results than all the other activation functions. Moreover, we can see that some improved ReLU functions performed slightly better or even poorer than the original ReLU function, implying that these improved ReLUs did not process the negative phase information in a proper way. 

\subsection{Channel Compensation}
We have mentioned that reborn mechanism is also a sort of channel compensation method. In this section, we designed some experiments to validate this. 

Specifically, we reduced the number of convolutional kernels by half in all convolutional layers, which actually means the number of learned parameters was reduced fourfold, and the ReLU functions in the original three basic CNN models are replaced with reborn mechanism as the nonlinear activation method. We also conducted the experiments about another classic channel compensation method, the CReLU function. Table~\ref{tab2} showed all the experimental results, where the mark Half means the number of feature channels was halved. From Table~\ref{tab2}, we can see that although the number of feature channels was halved, the model with reborn mechanism still obtained better results than the model which possessed a double number of channels with ReLU as the nonlinear activation method. This results demonstrated that reborn mechanism can be regarded as a channel compensation method and enhance the utilization efficiency of the feature channels. Therefore, the CNN model with reborn mechanism can use fewer convolutional kernels but achieve the same or even better performance. 
Compared to the CReLU which is another channel compensation method, our reborn mechanism obtained better results. We believe reborn mechanism utilizes the negative feature channels more properly and conducts well-designed operation on the negative channels. This is significantly different from and superior to the way of CReLU that concatenates negative channels with the positive ones straightforwardly.


\subsection{Further Analysis about Efficacy}
It is noted that that reborn block contains more learned parameters. To clarify that the more learned parameters is not the key reason for superiority of reborn mechanism, we designed some comparative experiments. Specifically, with the CNN models as the baselines, we still utilize ReLU as the nonlinear activation function, simply adding one $3\times3$ convolutional layer which has the same number of filters as the former $3\times3$ convolutional layer after each ReLU. These new added convolutional layers are followed by the next originally existing convolutional layers immediately, which means that there exist no nonlinear activation functions between these two convolutional layers. We call these original convolutional layers added with new convolutional layers as extended convolutional layers. Still, we exploited the same training methods. 

Table~\ref{tab3} reported the experimental results. It is observed that the difference between the standard method which uses ReLU simply and the extended convolutional layers is insignificant and negligible. Moreover, it is obvious that the negative feature maps screened out by ReLU function in reborn block are much sparser than the original feature maps, therefore the extended convolutional layers actually have more useful learned parameters than the reborn block. With this advantage, the extended convolutional layers still cannot improve the basic CNN models significantly. Thus we can conclude that more parameters do not necessarily improve the model representation ability and it might not be the key reason for the effectiveness of reborn mechanism. From another aspect, although extended convolutional layers contain more learned parameters, it does not have more nonlinear activation layers. Therefore, the representation ability of the model cannot be improved. 

Instead, we conjecture that three reasons may explain why our reborn mechanism works. First of all, reborn mechanism actually proposes a more proper way to utilize negative phase information effectively. Second, as discussed above, reborn block leads to a unsymmetrical CNN structure which can enhance further the model representation ability. Third, reborn block can be seen as a sort of channel compensation method able to promote the utilization efficiency of feature channels.

\subsection{Visualization}
Fig.~\ref{pic} visualizes various feature mappings activated by reborn mechanism and ReLU. To exhibit the difference more clearly, the CNN model we utilized is relatively small. It was composed of five convolutional layers, the $n$-th layer possessed $2^{n+4}$ channels. We extracted the feature mappings of the first convolutional layer, which is supposed to extract features like contours. It can be clearly observed that the feature mappings activated by ReLU lose much information and fail to detect distinct outlines compared to these activated by reborn mechanism. This phenomenon validates the effectiveness of our reborn mechanism.

\section{Conclusion}
In this paper, we innovate reborn mechanism to utilize the input information effectively, reduce loss of information flow, strengthen representation ability, and enhance the efficiency of filters. Moreover, we argue that reborn mechanism is not a simple nonlinear activation function but an innovative nonlinear activation mechanism.
We have detailed the novel reborn mechanism, interpreted our method from various perspectives, and conducted extensive experiments to validate its effectiveness. Experimental results showed that our reborn block can enhance the representation ability of CNN models significantly and lead to superior performance to those with ReLU or its improved versions. 




\end{document}